# A General Framework for Density Based Time Series Clustering Exploiting a Novel Admissible Pruning Strategy

Nurjahan Begum[1], Liudmila Ulanova[1], Hoang Anh Dau[1], Jun Wang[2], and Eamonn Keogh[1]

**Abstract**— Time Series Clustering is an important subroutine in many higher-level data mining analyses, including data editing for classifiers, summarization, and outlier detection. It is well known that for similarity search the superiority of Dynamic Time Warping (DTW) over Euclidean distance gradually diminishes as we consider ever larger datasets. However, as we shall show, the same is not true for clustering. Clustering time series under DTW remains a computationally expensive operation. In this work, we address this issue in two ways. We propose a novel pruning strategy that exploits both the upper and lower bounds to prune off a very large fraction of the expensive distance calculations. This pruning strategy is admissible and gives us provably identical results to the brute force algorithm, but is at least an order of magnitude faster. For datasets where even this level of speedup is inadequate, we show that we can use a simple heuristic to order the unavoidable calculations in a most-useful-first ordering, thus casting the clustering into an anytime framework. We demonstrate the utility of our ideas with both single and multidimensional case studies in the domains of astronomy, speech physiology, medicine and entomology. In addition, we show the generality of our clustering framework to other domains by efficiently obtaining semantically significant clusters in protein sequences using the Edit Distance, the discrete data analogue of DTW.

———————————— ◆ ————————————

## 1 INTRODUCTION

BECAUSE of the prevalence of time series data in human endeavors, the research community has made substantial efforts to create efficient algorithms for classification, clustering, rule discovery, and anomaly detection for this data type [1][4][23][35][49]. In particular, time series *clustering* is very useful, both as an exploratory technique and as a sub-module for solving higher-level data mining problems. As a motivating example, consider Fig. 1, which illustrates a subset of a cluster we discovered in a social media dataset [59]. This clustering allows us to at least partly address two problems:

- **Synonym Discovery:** In this example, we have a time series containing the volume of the hashtag *#Michael* over time. It is not clear to whom this refers: Michael Phelps? Michael Caine? However, by noting that this cluster also contains *#MichaelJackson*, this ambiguity is resolved.

- **Association Discovery:** Here we see that *#kanyewest* and *#taylorswift* have highly similar time series representations, but are clearly not synonyms. If we test to see whether this relationship existed prior to the illustrated timeframe, we find it does not. This suggests the existence of an *event* that caused this tempo-rary association, and with a little work we can discover that the famous "*I'mma let you finish*" event at the 2009 Video Music Awards produced this relationship [62].

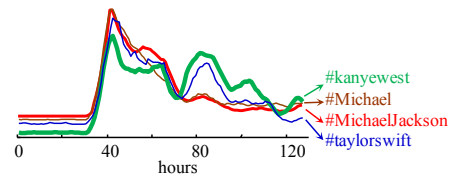

Fig. 1. A cluster of four Twitter hashtag usage time series (normalized for volume) over ~6 days starting from June 12, 2009 [59]. (Best viewed in color)

In this example, the knowledge gleaned is clearly trivial; however, similar ideas have been used to track the levels of disease activity and public concern during the recent influenza A H1N1 pandemic [48]. Note that while we discovered *this* example using Dynamic Time Warping (DTW), it might have been discovered more efficiently with the Euclidean distance (ED). However, in cases where there is a *causal relationship* (rather than just an *association*) between events, a local lag can result between peaks. It has been extensively shown in the literature that the 40-year old distance DTW is an ideal similarity measure to capture/be invariant to such out-of-sync relationships [47].

We argue that the problem we wish to solve, *robustly clustering large time series datasets with invariance to irrele-*

————————————————————
*1  Nurjahan Begum, Liudmila Ulanova, Hoang Anh Dau and Eamonn Keogh are affiliated with the University of California, Riverside, CA- 92521. E-mail: {nbegu001, lulan001, hdau001, eamonn}@cs.ucr.edu.*
*2  Jun Wang is affiliated with the University of Texas at Dallas, Richardson, TX - 75080. E-mail: wangjun@utdallas.edu.*

*vant data,* has not been solved before.

For most time series data mining algorithms, the quality of the output depends almost exclusively on the distance measure used [49]. In the last decade, a consensus has emerged that the DTW distance measure is the best measure in most domains, almost always outperforming the Euclidean Distance (ED) and other purported rivals [49]. As a concrete example, consider the two clusterings of three randomly chosen mammals shown in Fig. 2. The input data is the mitochondrial DNA after it was converted to a time series representation (converting DNA to time series is a commonly used operation [40][41]). Two types of DNA mutations, *insertions* and *deletions*, have the effect of "warping" the time series. At least in this case, we can see that DTW is invariant to these mutations and correctly unites *Bos taurus* (cattle) and *Hyperoodon ampullatus* (bottlenose whale), with *Talpa europaea* (mole) as the out-group.

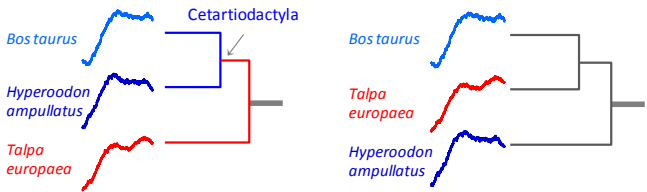

Fig. 2. Single-linkage hierarchical clusterings of DNA using DTW (*left*) and Euclidean distance (*right*).

While this example[1] is on a small and somewhat specialized dataset in the domain of bioinformatics, in Section 6 we will show that the superiority of DTW extends to large datasets in many domains including discrete protein data.

### 1.1 Why This Problem Is Hard

Because DTW is intrinsically slow due to its quadratic time complexity, there are two ideas that are commonly used to mitigate the problem of using such a sluggish distance measure [41]. We briefly discuss them here only to dismiss them as possible solutions.

- *The convergence of DTW and Euclidean distance results for increasing data sizes.* It has been noted that for many problems, including motif discovery [35] and classification [49], the results returned by DTW and Euclidean distance tend to become increasingly similar as the dataset sizes increase. This suggests that it is more efficient to use Euclidean distance to cluster large datasets.

- *The increasing effectiveness of lower-bounding pruning for increasing data sizes.* For some problems, notably similarity search, the lower-bounding pruning of unnecessary calculations is the main technique used to produce speedup. The effectiveness of this lower-bounding tends to improve dramatically as the datasets get larger [41].

Unfortunately, neither of these observations helps us for *clustering* under DTW. To demonstrate why the first observation does not help, we performed a simple experiment in which we measured the leave-one-out training error of 1-NN classification using both DTW and ED, for various numbers (50 to 2000) of exemplars from the CBF dataset [25]. With 50 objects, the error rates differ by a factor of 4.6 (7% and 1.75%, respectively), but as shown in Fig. 3.*top*, by the time we consider the 2,000 object dataset, this difference is essentially zero.

This effect is well known for time series *classification* [44][49], and it might be imagined that this also applies to *clustering*. To show that this is not the case, we performed a parallel experiment in which we *clustered* the same objects and measured the performance using the Rand Index [43]. As shown in Fig. 3.*bottom*, DTW *clustering* maintains its superiority over Euclidean distance as the datasets get larger.

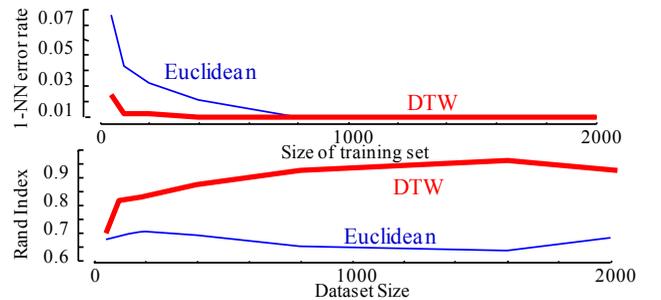

Fig. 3. *top*) The classification error rates of DTW and ED tend to converge as we see more training data. *bottom*) In contrast, for clustering, DTW retains its great superiority over ED for increasingly large datasets.

Similarly, the second observation above does not help significantly to prune DTW distance computation for clustering. It is true that lower bounds are increasingly effective for larger datasets when attempting a *similarity search*. This is because for larger datasets, we can expect to have a smaller *best-so-far* early on, which allows more effective pruning [41][44][49]. However, in *clustering*, we need to know the distance between all pairs [23], or at least all distances within a certain range, which renders the typical use of lower-bounding pruning ineffective.

### 1.2 Why Existing Work Is Not the Answer

More generally, many clustering algorithms achieve scalability by exploiting a spatial access method. For example, the scalable version of the ubiquitous DBSCAN uses an R*tree [12]. However, because DTW is not a metric, it is very difficult to index, especially for long (i.e., high-dimensional) time series objects.

Beyond the need to scalably support DTW, we note the need for a clustering algorithm that supports invariance to outliers. That is to say, unlike some clustering methods such as k-means, which attempt to explain *all* the data, we believe it is especially important to allow a time series clustering algorithm the freedom to ignore some data.

Consider the example in Fig. 4. We took twelve objects from a heraldic shield dataset [63] and clustered these using k-means and DP, the algorithm we propose to augment (described in detailed in Section 4). Because we are using the (non-metric) DTW measure, which may

---

prevent k-means from converging, we used the variant in [21] which performs k-means clustering using the all-pair distance matrix. Note that for the ease of visualization, we used multidimensional scaling to cast high-dimensional time series objects to two dimensions. After we ran the algorithms, both of them gave a perfect Rand Index score of 1.0. We then inserted a single outlier object (object 12) from this dataset and reran the algorithms. As we can see from Fig. 4. *bottom.left*), k-means assigned objects 6-11 to the cluster of the outlier object. In addition, k-means falsely identified objects 1 and 2 as a separate cluster from the cluster of objects 3-5. In contrast, from Fig. 4.*bottom.right*) we can see that DP only clustered object 7 in the cluster of the outlier object, but did not change the cluster labels of the rest of the dataset.

This toy example is contrived and anecdotal, but conformed by more rigorous and wide-reaching experiments on real data [64].

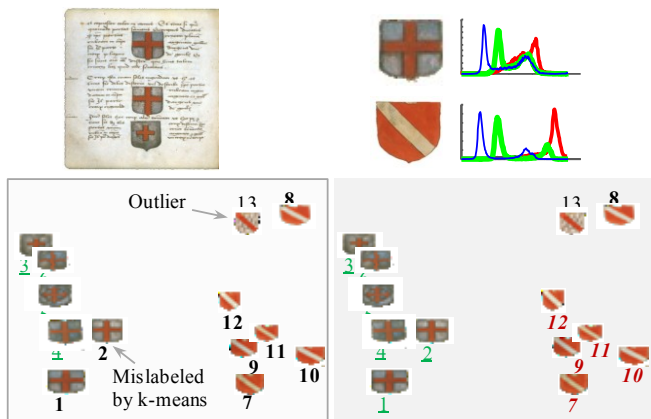

Fig. 4. *top.left*) Leaf 18V of a 15th-century book, *Treatises on Heraldry* [63]. *top.right*) The colorful heraldic shields can be converted to 3D RGB "time series" of color distribution. *bottom.left*) Even the insertion of a single outlier can confuse k-means. *bottom.right*) In contrast, the performance of the DP algorithm is not sensitive to outliers.

In this work, we address all the considerations above. We adapt DP (Density Peaks), a relatively new clustering framework that is able to ignore outlying data points [45]. While DP is insensitive to outliers, it is relatively slow, as it requires O($N^2$) DTW calculations. We augment DP such that it can exploit *both* DTW upper and lower bounds to compute only the absolutely necessary DTW calculations, and do so in a *best-first* manner, giving our algorithm the desirable *anytime algorithm* behavior [3][61].

## 2 RELATED WORK

The field of clustering is vast, and even the subfield of clustering time series has an enormous literature [1][23][57][60]. Much of the works on time series clustering are concerned with clustering based on time series *features* [57], which are at best tangentially related to our goals. Here, we are only interested in clustering based on time series *shapes*. In the latter case, there are two important and interrelated choices that define most of the literature: the choice of distance measure and the choice of clustering algorithm.

Most of the literature on time series shape-based clustering uses metric measures such as Euclidean distance [10][57]. The ubiquity of Euclidean distance seems to derive more from its familiarity and ease of indexing than any data-driven assessment of its effectiveness. As Fig. 3.*top* illustrates with a single representative example, the general superiority of DTW over ED is well understood in the community (cf. [49]), at least for *classification*. As 3.*bottom* suggests, and as we later empirically demonstrate on many diverse datasets, the dominance of DTW over ED for *clustering* is, if anything, much greater.

The plethora of shape-based clustering algorithms [23][42][60] can be divided at the highest level into those that insist on explaining (i.e., *clustering*) all the data [60] vs. those that have the representational power to leave some data unclustered (a small minority) [42]. We believe that this distinction is underappreciated and critical to the success of most efforts. For clarity, consider the following analogy: if we were clustering *people*, surely every person in our database would belong to some group, even if (due to the small size of our sample) the size of some groups were just *one*. In contrast, if we were clustering subsequences from a speech articulation database (see Section 6), we would hope that the subsequences would cluster into well-defined words or phrases. However, it is highly likely that we would have some examples of coughing, sneezing or harrumphing. Such sequences are likely to be very dissimilar to the rest of the database. Not only do we not want/need them to be clustered, but also we do not want them to affect the clustering of the clusterable words or phrases (recall Fig. 4). However k-means and its variants insist on explaining these instances and because of k-means's sum of squares objective function, these highly dissimilar items have a huge effect on the quality of the overall clustering.

One of the basic questions in clustering problems is the notion of a 'cluster' itself. There exist partitional clustering algorithms such as k-means which typically assume data has balanced Gaussian "ball-shaped" clusters. Whereas some other algorithms (e.g. DBSCAN [12]) take the *density* of objects into consideration regardless of the shape the clusters may have. The intuition behind DBSCAN algorithm is that, each object in a cluster must have at least some number of other objects (*MinPts*) in its vicinity (ε). If two objects are *density-reachable* from each other, then they are assigned in the same cluster. However, DBSCAN is not deterministic in its assignment of the cluster border points, because the result specifically depends on the order of the objects considered. Besides, DBSCAN has two parameters, *MinPts* and ε, and there is no concrete strategy for setting these parameters.

There exist works [33] in the literature that perform clustering on top of DBSCAN [12]. The problems with such approaches are the inheritance of the non-determinism of DBSCAN and the use of *only* lower bounds to prune expensive distance calculations. A variant of DBSCAN called IncrementalDBSCAN [13] has been demonstrated to perform much better than the brute-force re-clustering of newly added/removed objects which is mostly appropriate for datasets changing incre-mentally. DBCLASD [58] is a non-parametric grid-based clustering algorithm which considers the density of objects like DBSCAN to define clusters. It exploits a grid-based approach to find polygons of clusters. However, this algorithm cannot be applied to high-dimensional time series domains. DENCLUE [18] is an algorithm which combines both density-based and grid-based approaches to define cluster centers that has the local maximum of some density function. The cluster assignment of other objects in the datasets is done by a hill-climbing approach. There exist other types of clustering algorithms like hierarchical and model-based [17]. However, such algorithms are known to be quite slow and not appropriate for our context.

A handful of research efforts [60] have attempted to mitigate the slow performance of DTW clustering by casting it to an anytime framework. Most such efforts reduce to the following: until there is a user interrupt, these frameworks keep replacing the (fast to compute) approximate DTW distances with true (slow to compute) DTW distances. If there is no user interrupt, such frameworks will calculate the full distance matrix (generally in some "*most-likely-to-be-useful*" order) and return the exact clustering. We refer readers interested in time series clustering to the detailed surveys [26][30].

Our proposed algorithm goes beyond this literature in several ways. Most importantly, we show that calculating the full distance matrix is unnecessary in the general case. By exploiting both upper and lower bounds to DTW, and, more critically, by exploiting the *relationship* between these bounds, we can compute the *exact* clustering while only calculating a tiny fraction of the full distance matrix.

## 3 BACKGROUND

There has been significant research on clustering datasets that are too large to fit in main memory [5][6]. This problem setting typically assumes inexpensive distance measures, but costly disk accesses [6]. However, the problem we wish to solve exploits DTW, which itself is a very expensive distance measure. Therefore, in situations when even the data can be stored in main memory, the time needed to do the clustering may be on the order of days/weeks. The problem we are interested in is therefore, CPU constrained, not I/O constrained.

### 3.1 Anytime Algorithms

For most clustering algorithms, it is well known that not all distance measurements contribute equally to the final clustering assignments. For example, a recent paper on hierarchical clustering demonstrates (under some mild assumptions) that it is possible to capture the true structure of the clustering with just *carefully chosen* $O(n \log^2 n)$ distance computations [28]. The fact that some distance computations are more important than others, immediately suggests that we should use *anytime algorithms*, assuming only that we can find an efficient and (even *somewhat*) effective test to identify these influential distance computations.

Recall that anytime algorithms are algorithms that can return a valid solution to a problem, even if interrupted before ending [60][61]. Starting with a negligibly small amount of setup time, these algorithms always have a *best-so-far* answer available and the quality of the answer improves with the increase of execution time. The desirable properties of anytime algorithms are interruptibility, monotonicity, measurable quality, diminishing returns, preemptibility, and low overhead [61]. Note that this is a very brief introduction to anytime algorithms; we refer the interested reader to [61], which contains an excellent survey. As Fig. 5 shows, anytime algorithms, in essence, optimize the tradeoff between execution time and quality of the solution.

For clarity, we reiterate that the anytime algorithm approach is just one of the two contributions of this work. We propose to make the clustering *absolutely faster* by admissible pruning. This is *in addition* to rearranging the order in which the non-prunable calculations are considered to produce the best possible diminishing returns *anytime algorithm* behavior.

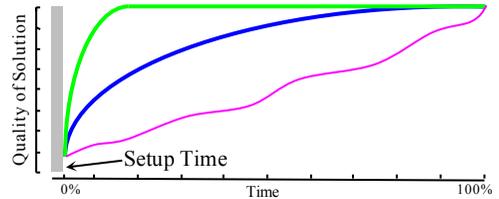

Fig. 5. An abstract illustration of an anytime algorithm. The three curves show a comparison of the possible performances of three hypothetical anytime algorithms. The bottommost curve (pink) is only improving linearly over time, but the topmost curve (green) demonstrates *diminishing returns*, making most of its improvements early on.

## 4 ALGORITHM

We begin by reviewing the basic algorithm that we will augment and specialize to handle DTW.

### 4.1 Density Peaks Algorithm Overview

Our proposed solution is inspired by DP, the density-based clustering algorithm recently proposed in [45]. We chose to augment the DP framework for solving large time series clustering problems because of the following:

• Recent literature [42] and our own experience on real datasets (cf. Section 6) suggest that the successful clustering of time series requires the ability to ignore some data objects. The issue is not merely that anomalous objects themselves are unclusterable; but that the presence of these objects can affect the labels of objects that *are* clusterable in unpredictable ways. The DP algorithm has been shown to be able to ignore anomalous data points.

• The DP algorithm is able to handle datasets whose clusters can form arbitrary shapes. This is in contrast to k-means and related algorithms, which assume the clusters are "balls" in space. This observation is particularly important for DTW, which is not a metric. While we cannot exactly visualize DTW clusters in a metric space, it is clear that some classes of objects under DTW form complex manifolds in DTW "space."

- Many clustering algorithms require the user to set many parameters. In contrast, the DP algorithm requires only two. These parameters are relatively intuitive and not particularly sensitive to user choice.
- Finally, it happens to be the case that the DP algorithm is amiable to optimization and conversion to an anytime algorithm.

To make our argument more concrete, we will take the time to explain the clustering algorithm [45] we adapt and augment. The DP algorithm assumes that the cluster centers are surrounded by lower local density neighbors and are at a relatively higher distance from any point with a higher local density. Therefore, for each point $i$ in the dataset, the DP algorithm computes two quantities:

- Local density $(\rho_i)$
- Distance from points with higher local density $(\delta_i)$.

We can formally define these two quantities:

**Definition 1** The *Local Density* $\rho_i$ of point $i$ is the number of points that are closer to it than some cutoff distance $d_c$.

**Definition 2** The *Distance from Points of Higher Density* is the minimum distance $\delta_i$ from point $i$ to all the points of higher density. For the special case of the highest density point, this distance is the maximum of the distances of all the points from their higher density points.

We give the algorithm to compute $\rho_i$ in TABLE 1 and $\delta_i$ in TABLE 2.

TABLE 1

LOCAL DENSITY CALCULATION ALGORITHM

| Input: | **D**,all-pair distance matrix |
| | **d$_c$**, cutoff distance |
| Output: | **ρ**,the local density vector for all n points in the dataset |
| 1 | **for** i = 1:n |
| 2 | ρ(i) = count(D(i,otherObjects)<d$_c$) |
| 3 | **end** |

Given the all-pair distance matrix $D$ and a cutoff distance $d_c$, for each point $i$ in the dataset, $\rho_i$ is calculated in lines 1-3 of TABLE 1. In TABLE 2, using the local densities $\rho$ from TABLE 1, for each point $i$, the list of the points with higher densities is calculated (line 2). In line 4, this list is sorted in descending order. From lines $5 - 7$, for each point in the sorted order, the distances from their higher density points are calculated. For the special case of the highest density point (which by definition has no higher density neighbor), this distance is calculated in line 8.

Given the $\rho_i$ and $\delta_i$ for each object $i$, the DP algorithm calculates the cluster centers $\chi$, and performs the cluster assignments based on these centers.

TABLE 2

DISTANCE TO HIGHER DENSITY POINTS ALGORITHM

| Input: | **D**,all-pair distance matrix |
| | **ρ**,the local density vector |
| Output: | **δ**,NN distance vector of higher density points |
| 1 | **for** i = 1:n |

| 2 | δ_list(i) = findHigherDensityItems(i,ρ) |
| 3 | **end** |
| 4 | [sorted_δ_list, sortIndex] = sort(δ_list,'descend') |
| 5 | **for** j = 2:n |
| 6 | δ(sortIndex(j)) = NNDist(sorted_δ_list(j)) |
| 7 | **end** |
| 8 | δ(sortIndex(1)) = max(δ(2:n)) |

The cluster centers are selected using a simple heuristic: *points with higher values of* $(\rho_i \times \delta_i)$ *are more likely to be centers*. We give the cluster center selection algorithm in TABLE 3.

TABLE 3

CLUSTER CENTER SELECTION ALGORITHM

| Input: | **δ**,NN distance vector of higher density points |
| | **ρ**,the local density vector |
| | **k**, number of clusters |
| Output: | **χ**, cluster centers |
| 1 | χ = topK(sort(ρ*.δ, 'descend'),k) |

Given the sorted values of $(\rho_i \times \delta_i)$ in descending order, the top $k$ items are selected as cluster centers (line 1). The value of $k$ can be specified by the user, or found automatically using a "knee-finding" type of algorithm [45].

The final step of the DP algorithm is the cluster assignment. Each data item gets the cluster label of its nearest neighbor (NN) from the list of points with higher local densities than it has. We give the cluster assignment algorithm in TABLE 4.

TABLE 4

CLUSTER ASSIGNMENT ALGORITHM

| Input: | **χ**, cluster centers |
| | **δ**,NN distance vector of higher density points |
| | **sortIndex**, sorted index of items based on descending ρ |
| Output: | **C**, clusters |
| 1 | **for** i = 1:size(χ) |
| 2 | C(χ(i)) = i //assign cluster labels for centers |
| 3 | **end** |
| 4 | **for** j = 1:n |
| 5 | **if** C(sortIndex(j)) == empty //no cluster label yet |
| 6 | C(sortIndex(j)) = C(NN(sortIndex(j))) |
| 7 | **end if** |
| 8 | **end** |

In lines 1-3 the cluster labels of the centers are assigned. After this initialization, each of the points in the dataset (other than the centers themselves) gets the cluster label of its nearest neighbor from the higher density list in the descending order of local density (lines 4-8). It is important to note that this algorithm allows the clusters to have arbitrary, possibly non-convex shapes, unlike k-means and its variants, which are restricted to a Voronoi partitioning of the input space. We are now in a position to describe our augmented version of the DP framework.

## 4.2 TADPole: Our Proposed Algorithm

We call our algorithm, TADPole (Time-series Anytime DP). As stated in Section 1, in order for the original DP algorithm to cluster a dataset, we need to know the distances between all pairs. The time needed to compute these all-pair distances becomes untenable for a quadratic time distance measure such as DTW. In order to mitigate this undesirable time complexity, our thoughts naturally turn to attempts to speed up other (non-clustering) algorithms that need to compute DTW frequently. Most such algorithms exploit linear time lower bounds like LB_Keogh [24], LB_Kim, LB_Yi [57], etc. Moreover, some algorithms exploit the fact that ED is an *upper* bound to DTW, and can also be computed in O($n$) time.

In our TADPole algorithm, we augment the DP clustering framework and exploit the upper and lower bounds of DTW to prune unnecessary distance computations, which results in at least an order of magnitude speedup. For datasets where even this level of speedup is inadequate, we show that we can use a simple heuristic to order the unavoidable calculations in a most-useful-first ordering. As a result, our algorithm can be cast as an *anytime* clustering framework, which quickly produces a good answer and rapidly refines it until it converges to the exact answer (for proof, c.f. Section 10).

The inputs to the TADPole algorithm are the lower bound and upper bound matrices for the true DTW distances of all the objects of the dataset. Note that the time needed to compute these is inconsequential (<1%) relative to the overall clustering time.

The only parameters we need are the cutoff distance ($d_c$) and optionally, the number of clusters ($k$), if the user wishes to specify this value rather than use the knee-finding heuristic suggested in [45]. Because we use DTW as the underlying distance measure, the warping window size is another parameter for TADPole. We discuss the heuristic to set these parameters in Section 5.3. Note that our use of two additional upper bound and lower bound matrices increases the space complexity of the algorithm by 200%. However, this is not an issue because:

- The DP algorithm (especially when using DTW or another expensive measure) is *CPU* bound, not *space* bound.

- If really necessary, we could greatly mitigate this space overhead. The lower bound matrix will have many elements that are zeros, and thus would be amiable to be encoded as a sparse matrix.

- If needed, both the upper and lower bound matrix's could be computed in a *just-in-time* fashion [41]. This would greatly reduce the memory footprint, at the expense of a more complicated implementation.

For clarity of presentation, we present our contributions in two different sections, although the final algorithm incorporates both ideas. In Sections 4.2.1 to 4.2.4, we show how to accelerate the TADPole algorithm by admissibly pruning the distance computations during the calculation of local densities ($\rho$) and NN distances ($\delta$) from a higher density list for each item. In Section 4.2.5, we show how to reorder these computations to give us the *diminishing returns* property of anytime algorithms

[3][61].

### 4.2.1 Pruning During Local Density Calculation

Consider the four cases shown in Fig. 6.

In this step of the TADPole algorithm, the inputs are the fully computed lower (LB$_{Matrix}$) and upper bound (UB$_{Matrix}$) matrix. For each object pair ($i,j$), while calculating their local densities (lines 1 to 3 in TABLE 1), we prune their distance ($D_{ij}$) computation according to the following four cases shown in Fig. 6:

**Case A:** *Objects $i$ and $j$ are identical*

The DTW distance of two identical objects, $i$ and $j$, is equal to their ED distance. It is a simple lookup in the upper bound distance matrix, and requires no actual DTW distance computation. This case is logically possible but very rare (Fig. 6.A)).

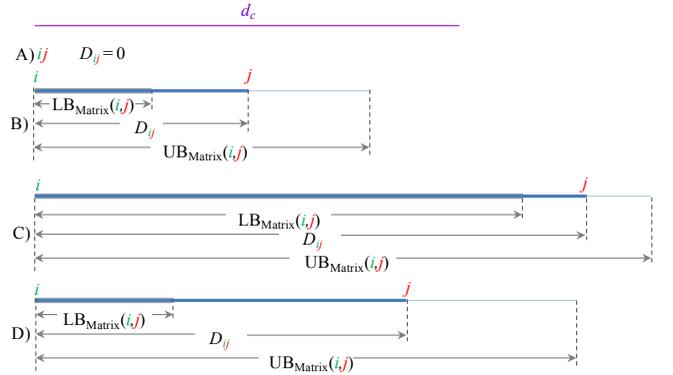

Fig. 6. The four mutually exclusive and exhaustive cases of distance computation pruning during local density calculation. Note that the cutoff distance $d_c$, represented by the purple line at the top, applies to all four cases below it. Case A is difficult to visually represent, as $i$ and $j$ coincide.

**Case B:** UB$_{Matrix}$($i,j$) < $d_c$

If the upper bound distance between objects $i$ and $j$ is less than the cutoff distance ($d_c$), then $i$ and $j$ are definitely within $d_c$ distance to each other (Fig. 6.B)). Therefore, we can prune the DTW distance computation of these two objects.

**Case C:** LB$_{Matrix}$($i,j$) > $d_c$

If the lower bound distance of $i$ and $j$ is greater than the cutoff distance, then these two objects are definitely *not* within $d_c$ distance to each other (Fig. 6.C)). We can therefore admissibly prune their DTW distance computation.

**Case D:** LB$_{Matrix}$($i,j$) < $d_c$ and UB$_{Matrix}$($i,j$) > $d_c$

In this case, we cannot tell whether or not the actual DTW distance between $i$ and $j$ is within $d_c$. Therefore, *only in this case* do we need to compute $D_{ij}$ (Fig. 6.D)).

With this intuition in mind, we specify the formal distance pruning algorithm during the local density calculation in TABLE 5.

As we can see from TABLE 5, in lines 5 - 21, for all the object pairs in the data, the TADPole algorithm checks which of the four cases applies in order to determine whether or not these objects are within the cutoff distance.

The occurrence of case B tells us that the object pair in question is definitely within $d_c$ (lines 10 -11) without having to calculate the expensive true DTW distance. Cases A (lines 8 -9) and C (lines 12 -13) specify that the object pair is not within $d_c$. It is only the occurrence of case D that forces the algorithm to calculate the true DTW distance of the object pair in question (lines 14 -19).

At the end of this section of TADPole, for each object $i$, we have all the local densities ($\rho_i$) computed. Using lines 1 - 3 of TABLE 2, we can now find the δ list, the list of the points with higher densities. Next we will describe our pruning strategy for this step.

TABLE 5

PRUNING ALGORITHM DURING LOCAL DENSITY CALCULATION

| | |
|---|---|
| Input: | **LB$_{Matrix}$**, full computed lower bound matrix |
| | **UB$_{Matrix}$**, full computed upper bound matrix |
| | **Data**, the dataset |
| | **d$_c$**, cutoff distance |
| Output: | **ρ**,local density vector for all points in dataset |
| | **D$_{Sparse}$**, partially filled distance matrix |
| 1 | D$_{Sparse}$ = empty |
| 2 | **for** i = 1:**size**(Data) |
| 3 | objectsWithin_dc = empty |
| 4 | **for** j = 1: **size**(Data) |
| 5 | **if** i == j |
| 6 | continue; |
| 7 | **else** |
| 8 | **if** LB$_{Matrix}$(i,j) == UB$_{Matrix}$(i,j) //case A) |
| 9 | continue |
| 10 | **elseif** UB$_{Matrix}$(i,j) < d$_c$ //case B) |
| 11 | objectsWithin_dc = [objectsWithin_dc j] |
| 12 | **elseif** LB$_{Matrix}$(i,j) > d$_c$ // case C) |
| 13 | continue |
| 14 | //case D) |
| 15 | **elseif** LB$_{Matrix}$(i,j) < d$_c$ **and** UB$_{Matrix}$(i,j) > d$_c$ |
| 16 | D$_{Sparse}$(i,j) = calculateDist(Data(i),Data(j)) |
| 17 | **if** D$_{Sparse}$(i,j) < dc |
| 18 | objectsWithin_dc = [objectsWithin_dc j] |
| 19 | **end if** |
| 20 | **end if** |
| 21 | **end if** |
| 22 | **end for** |
| 23 | ρ(i) = **length**(objectsWithin_dc) |
| 24 | **end for** |

### 4.2.2 Pruning During NN Distance Calculation from Higher Density List

Our pruning strategy for this step works in two phases. First, for each item we find an upper bound of the NN distance from its higher density list. In the second phase we perform the actual pruning based on these upper bounds. The distance computation of TADPole terminates when for all objects in the dataset, we are done finding their actual NN distance from their respective higher density lists.

**Phase 1:** Upper bound calculation

Given D$_{Sparse}$ and $\rho_i$ for each item $i$, we initialize the upper bound of its NN distance from its higher density list, $ub_i$, to inf. For each item $j$ in the higher density list of $i$, we either have the actual DTW distance ($D_{ij}$) computed already or have access to the upper bound (UB$_{Matrix}$(i,j)) to this distance. We scan the higher density list of item $i$, and if the current $ub_i > D_{ij}$ or $ub_i >$ UB$_{Matrix}$(i,j), we update the current $ub_i$ to D$_{ij}$ (if available already), or to UB$_{Matrix}$(i,j) otherwise. Therefore, we can guarantee that the NN distance from the higher density list for item $i$ can be no larger than $ub_i$. We give a visual illustration of this upper bound calculation in Fig. 7.

In Fig. 7, the elements on object $i$'s higher density list are $j_1$ – $j_4$. Assume that we only know the DTW distances from object $i$ to objects $j_1$ and $j_3$, ($D_1$ and $D_3$ respectively, shown in blue). Because we do not know $D_2$ and $D_4$, we have shown these distances in gray in Fig. 7. When Phase 1 starts, $ub_i$ is initialized to inf. Now our TADPole algorithm scans object $j_1$ and updates $ub_i$ to $D_1$. Because UB$_{Matrix}$(i,$j_2$) and $D_3$ are both are greater than $ub_i$, we do not need to update $ub_i$. In the last step, given that UB$_{Matrix}$(i,$j_4$) < $ub_i$, we update $ub_i$ to UB$_{Matrix}$(i,$j_4$). This $ub_i$ is an upper bound of the NN distance from object $i$'s higher density list.

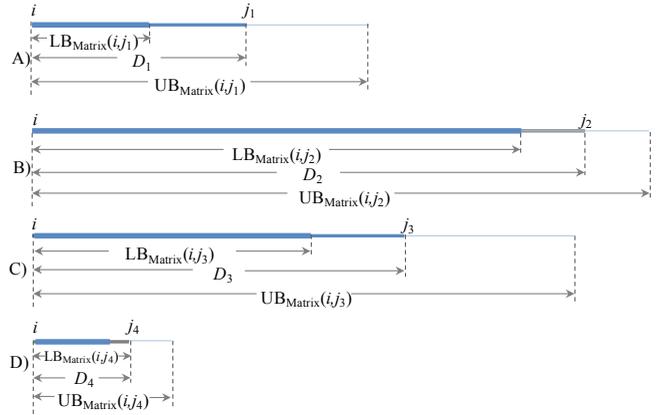

Fig. 7. An illustration of the distance pruning during the NN distance calculation from a higher density list of an object. From object $i$, the elements in the higher density list are $j_1$ – $j_4$. After Phase 1, $ub_i$ will be UB$_{Matrix}$(i,$j_4$). In Phase 2, the distance computations of $D_{ij2}$ and $D_{ij3}$ are pruned.

We give the upper bound calculation algorithm for the NN distance computation from a higher density list in TABLE 6. We initialize the upper bound vectors of NN distances of objects from their higher density list, $ub$ to inf (line 1). Next, considering each of the item on the higher density list of an object $i$, $\delta\_list_i(j)$, we check whether $i$'s current upper bound can be tightened (lines 5 -13). In lines 5 - 8 we see if the actual distance between $i$ and $\delta\_list_i(j)$ has been computed already, then whether or not this distance can tighten $ub_i$. If the distance has not been computed yet, then in lines 10 - 12 we check whether we can tighten $ub_i$ by replacing it with the upper bound distance between $i$ and $\delta\_list_i(j)$.

At the end of this phase of TADPole, we have $ub$, the upper bound vector of NN distances from higher density points, computed. We now describe exploiting $ub$ to prune the distance calculations during the computation of the higher density list.

TABLE 6

UPPER BOUND CALCULATION ALGORITHM FOR NN DISTANCE COMPUTATION FROM HIGHER DENSITY LIST

| Input: | $UB_{Matrix}$, full computed upper bound matrix |
| | **Data**, the dataset |
| | $D_{Sparse}$, partially filled distance matrix |
| | **δ_list**, list of the points with higher densities |
| Output: | *ub*, upper bound vector of NN distances from higher density points |
| 1 | $ub$ = inf(size(Data)) |
| 2 | **for** i = 1:size(Data) |
| 3 | **for** j = 1:size(δ_list$_i$) |
| 4 | highDensityItem = δ_list$_i$(j) |
| 5 | **if** $D_{Sparse}$(i, highDensityItem)≠ empty |
| 6 | **if** $ub_i$ > $D_{Sparse}$(i, highDensityItem) |
| 7 | $ub_i$ = $D_{Sparse}$(i, highDensityItem) |
| 8 | **end if** |
| 9 | **else** |
| 10 | **if** $ub_i$ > $UB_{Matrix}$(i,highDensityItem) |
| 11 | $ub_i$ = $UB_{Matrix}$(i,highDensityItem) |
| 12 | **end if** |
| 13 | **end if** |
| 14 | **end for** |
| 15 | **end for** |

**Phase 2:** Pruning

We give the pruning algorithm during the computation of NN distances from the higher density lists of all objects in TABLE 7.

We begin by scanning the higher density list of each of the objects again. In line 5 of TABLE 7, for an object *i*, we test whether $LB_{Matrix}(i, δ\_list_i(j))$ is greater than $ub_i$ we calculated in TABLE 6 . If this is the case, we prune the distance computation (line 6) for δ_list$_i$(j). Otherwise, if the *true* distance between i and δ_list$_i$(j) is already calculated, then we consider this distance as one of the potential NN distances from *i's* higher density list (line 9). If the true distance is not yet known, *only then* do we compute it (line 11-12). Finally, we compute the NN distance vector for all objects from their higher density lists (line 17).

In Fig. 7, we see that both $LB_{Matrix}(i,j_2)$ and $LB_{Matrix}(i,j_3)$ are greater than $ub_i$. Therefore, we can prune $D_2$ and $D_3$. In this example, we assumed we know $D_1$; therefore, after the pruning done by Phase 2, we only need to calculate $D_4$.

TABLE 7

PRUNING ALGORITHM DURING THE COMPUTATION OF THE NN DISTANCES FROM THE HIGHER DENSITY LISTS OF ALL OBJECTS

| Input: | $LB_{Matrix}$, full computed lower bound matrix |
| | **Data**, the dataset |
| | $D_{Sparse}$, partially filled distance matrix |
| | **δ_list**, list of the points with higher densities |
| | *ub*, upper bound vector of NN distances from higher density points |

| Output: | **δ**, NN distance vector of higher density points |
|---|---|
| 1 | **for** i = 1:size(Data) |
| 2 | temp_δ = empty |
| 3 | **for** j = 1:size(δ_list$_i$) |
| 4 | highDensityItem = δ_list$_i$(j) |
| 5 | **if** $LB_{Matrix}$(i,highDensityItem)> $ub_i$ |
| 6 | **continue** //prune distance computation |
| 7 | **else** |
| 8 | **if** $D_{Sparse}$(i, highDensityItem)≠ empty |
| 9 | temp_δ = [temp_δ $D_{Sparse}$(i, highDensityItem)] |
| 10 | **else**   // calculate distance |
| 11 | $D_{Sparse}$(i, highDensityItem) = |
| 12 | calculateDist(Data(i),Data(highDensityItem)) |
| 13 | temp_δ = [temp_δ $D_{Sparse}$(i, highDensityItem)] |
| 14 | **end if** |
| 15 | **end if** |
| 16 | **end for** |
| 17 | δ(i) = min(temp_δ) |
| 18 | **end for** |

After this phase of TADPole, for each item *i* we have access to the NN distance from points with higher local densities ($δ_i$). At this point, given $\varrho_i$ and $δ_i$ for each object *i*, the TADPole algorithm calculates the cluster centers χ using the algorithm in TABLE 3, and performs the cluster assignments based on these centers according to the algorithm in TABLE 4.

### 4.2.3   Multidimensional Time Series Clustering

While most of the research efforts on time series clustering have considered only the single-dimensional case [23][42], the increasing prevalence of medical sensors (c.f. Section 6.2) and wearable devices (c.f. Section 6) has given urgency to the need to support multidimensional clustering [50]. Fortunately, our extension of TADPole to the multidimensional case requires changing only a *single* line of code. For clarity, we highlight these changes for multidimensional clustering for TABLE 5 and TABLE 7 in TABLE 8 and TABLE 9, respectively.

TABLE 8

PRUNING ALGORITHM DURING LOCAL DENSITY CALCULATION FOR MULTIDIMENSIONAL DATA (SEE TABLE 5)

| Input: | $LB_{Matrix}$, full computed lower bound matrix |
| | along *d* dimensions |
| | $UB_{Matrix}$, full computed upper bound matrix |
| | along *d* dimensions |
| | Data, the dataset |
| | $d_c$, cutoff distance |
| Output: | **δ**, NN distance vector of higher density points |
| | ... |
| 16 | $D_{Sparse}$(i,j) = $\sum_{dim=1}^{d}$ calculateDist(Data$_{dim}$(i), Data$_{dim}$(j)) |
| | ... |

| 10 | **else**    // calculate distance |
| 11 | $D_{Sparse}(i, highDensityItem) =$ |
| 12 | $\sum_{dim=1}^{d} calculateDist(Data_{dim}(i), Data_{dim}(highDensityItem))$ |

Recall that in TABLE 5, we gave the full lower bound and upper bound distance matrices as inputs to the algorithm. To perform multidimensional clustering, for each dimension we wish to consider, we calculate the corresponding lower bound/upper bound distance matrices *independently* along those dimensions. We take the sum of all lower bound matrices/upper bound matrices and give these cumulative matrices as inputs to our algorithm described in TABLE 5. In addition, when we actually calculate the distances (line 16 in TABLE 5 and lines 10-12 in TABLE 7), we take the summation of the distances along all the dimensions. All other components of TADPole will remain the unchanged. We explicitly evaluate the utility of TADPole clustering for multidimensional clustering in Section 6.4. More generally, as we shall show empirically in Section 5, by using the methods described above, TADPole can obtain *at least* an order of magnitude speedup over the original DP algorithm while producing identical results.

### 4.2.4  How Effective Is Our Pruning?

Here we will demonstrate *just* the utility of our pruning strategy, before generalizing to allow anytime behavior in the next section. In order to intuitively calibrate the effectiveness of our pruning, we compare TADPole to the best and worst possible cases of DP:

• In order to perform clustering, the DP algorithm needs the all-pair distance matrix computed [45]. Therefore, in terms of distance computation, the brute force DP algorithm itself is the obvious worst-case straw man.

• The best possible variant of DP is the one that performs a distance computation *only* when it is necessary. Therefore, during density computation, this variant of DP considers *only* those distance computations that contribute to the actual density of an object. In addition to this, during the computation of the NN distance from the higher density list of an object, this variant considers *only* the actual NN distances. We call this algorithm the *oracle* variant of DP. Note that we obviously cannot compute this in real-time, but only by doing an expensive post-hoc study.

We compare the amount of distance pruning we achieve against these two variants of the DP algorithm. For this experiment, we consider the StarLightCurves dataset [25]. We vary the number of objects in the dataset we need to cluster (by random sampling) and record the number of *true* DTW distance computations. As we can see from Fig. 8.*left*), the number of distance computations increases quadratically using the brute force algorithm. In contrast, the oracle algorithm requires very few distance computations; moreover, we can see that our TADPole algorithm performs *almost* as well as the oracle algorithm.

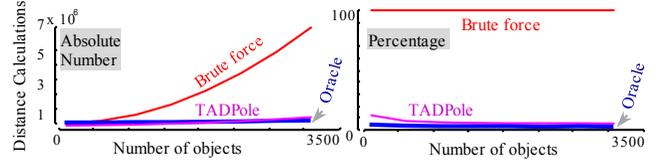

Fig. 8. A comparison of the amount of pruning TADPole achieved compared to an oracle and the brute force algorithm. TADPole is very close to the oracle algorithm.

This claim is reinforced in Fig. 8.*right*, in which we see that the *percentage* of distance computations TADPole requires is very close to the oracle algorithm. As we can see, as the datasets get larger, TADPole converges closer and closer to the oracle algorithm.

Also note that a similar performance is observed in *all* datasets we considered (archived in [64] for brevity). Moreover, we obtain similar results if we measure the CPU time instead. As we can see from Fig. 9, to cluster the StarLightCurves data, TADPole requires only 9 minutes, whereas the DP algorithm needs 9 hours.

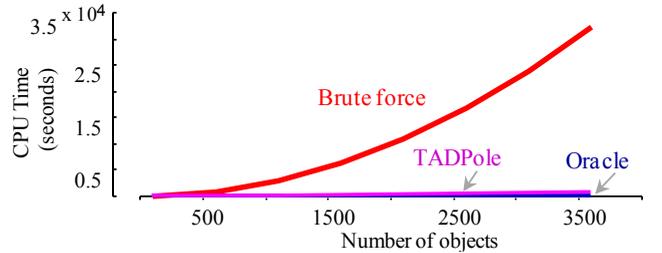

Fig. 9. A comparison of the CPU time spent by TADPole against an oracle and the brute force algorithm to cluster the StarLightCurves dataset. As before, the performance of TADPole is very close to the oracle algorithm.

To put these results into context we consider the results in a very recent research effort. In [33], the authors discuss the scalability of their clustering algorithm saying that "*For a large time series dataset with 9,236 objects with the length of each object n = 8192, it costs only 2.1 seconds to transform the whole dataset and an hour for clustering with DBSCAN*". Using this as a benchmark, we did the following experiment: we took 9,236 random walks each of length 8,192, and clustered this dataset using DP algorithm. DP took only 16 seconds to cluster this dataset. Of course the hardware settings of these two experiments may not be exactly commensurate, but it is clear we have lost nothing in terms of scalability by considering adopting the DP algorithm rather than the near ubiquitous DBSCAN [33]. As we will later show, we have also lost nothing in terms of accuracy.

In spite of these very promising results, which demonstrate a sixty-fold speedup, there exist datasets where even this amount of speedup is not adequate. In order to address similar scalability issues for other types of data mining analyses, including classification [52] and outlier detection [3][4], researchers have attempted to create *anytime* versions of their algorithms [3][60]. One significant advantage of the DP algorithm (and our modifications to it) is that it is particularly amiable to casting as an anytime framework. In essence, the computations discussed in this section can be computed in any order. Thus far, we have simply computed them in a top-to-bottom, left-to-right order. However, we should expect that not all such computations of the true DTW are equally significant in terms of their impact on the final clustering, and that if we could find even an approximate "most-significant-first" order, we could converge more quickly. In the next section, we describe such an ordering heuristic.

### 4.2.5 Distance-Computation Ordering Heuristic

Recall from the above that the DP algorithm may be considered a two-step algorithm; calculating the local densities (TABLE 1) first, and then finding the NN distances from the higher density lists (TABLE 2) of the objects. Only the latter step is amiable to anytime ordering; the former step may be regarded as the *setup time* [3][52][60].

Before attempting to create an anytime ordering function, it will be instructive to consider two baselines: what is the *best* we could possibly do, and what would we have to do in order to claim we are beating the most obvious straw men?

- The *best* ordering heuristic we consider is an *oracle* ordering. We can compute this by allowing the algorithm to *cheat*. In each step of the algorithm, this order chooses the object that maximizes the current Rand Index. The algorithm is cheating, because by definition, a clustering algorithm normally does not have access to class labels.
- The most basic straw man is top-to-bottom, left-to-right ordering, but that is brittle to "luck". A *random* ordering is much less so, so we consider random ordering as the baseline we would like to improve upon.

To understand the performance of these two algorithms, we took the Insect dataset from [25] and measured the Rand Index as the two algorithms above kept refining the mixture of true DTW distances and upper bound distances that we have at the end of phase 1 (i.e., TABLE 6), into the set of all *necessary* DTW computations needed (i.e., TABLE 7). The results are shown in Fig. 10 (for the moment, ignore the blue line). For completeness, we also show the accuracy achieved using the Euclidean distance with the DP algorithm. If the Euclidean distance *was* competitive, it would be fruitless to waste time computing expensive DTW calculations. The results clearly show that in *this* dataset, DTW is needed.

Being initially worse, the random ordering linearly (in a stepwise fashion) converges on the true clustering. In contrast, the oracle algorithm achieves a perfect Rand Index after calculating the true DTW distances for the NN list of just five objects.

As impressive as the oracle's performance is, we can actually come *very* close to it, as shown by the blue curve in Fig. 10. The ordering heuristic TADPole exploits is the descending order of the local density ($\varrho$) × the upper bound distance (*ub*) from the higher density list of the objects.

With a little introspection, it is easy to see why our distance computation-ordering heuristic is as close as the *oracle* ordering. Recall from Section 4 that points with higher values of $\varrho \times \delta$ are more likely to become cluster centers. Until we calculate all the NN from each object's higher density list, we do not have access to their $\delta$. However, we can estimate $\delta$ by the tight *upper* bound *ub* to $\delta$. Our distance computation-order heuristic exploits *ub* to prioritize distance computations for items that are more likely to be centers. Because the centers are selected earlier, we achieve a higher Rand Index with very few actual distance computations.

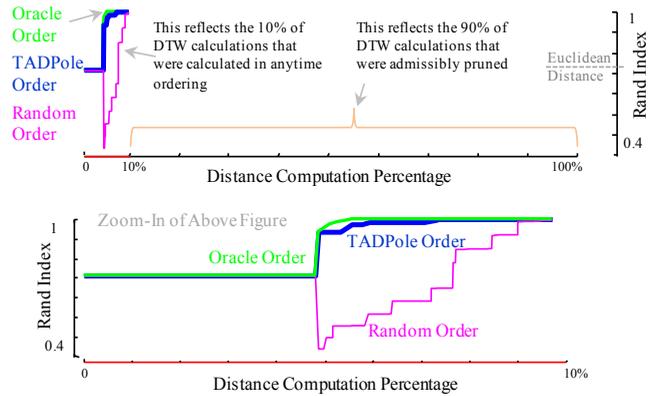

Fig. 10. *top*) A comparison of different distance computation-order heuristics on the Insect dataset [25]. An oracle ordering (green) converges stunningly quickly. The random ordering (pink) converges very slowly. Our proposed ordering (blue) is very close to the oracle. *bottom*) A zoomed-in view of the figure shown at the *top*).

From Fig. 10.*top*) we see that *in conjunction* with all the pruning strategies described in Sections 4.2.1 and 4.2.2, TADPole achieves a perfect Rand Index after doing *only* ~6% of all possible distance computations. Of course, it does not "realize" this, and must compute ~10% of all possible distance computations before admissibly terminating.

## 5 EXPERIMENTAL EVALUATION

All experiments in this paper (including the ones "inline" in the above text) are completely reproducible. We have archived all experimental code, parameter settings and data at [64]. The goal of our experiments is to show that our algorithm is more efficient and effective than current algorithms. We also show that our algorithm is not particularly sensitive to the *few* parameter choices we have to make. In addition to this, we demonstrate the utility of our approach on multiple real-world case studies.

We conducted our experiments on a Windows 8 machine with 3.5 GHz AMD A8-6500 APU with Radeon(tm) HD Graphics processor and 16 GB RAM. All our implementations are single threaded and were written in Matlab 7.9.0.529 (R2009b).

## 5.1 Comparison with State-of-the-Art Clustering Algorithms

The principle straw man we need to compare TADPole to is the brute force version of DP with DTW. This comparison is explicitly encoded in Fig. 10 and the similar figures below. In these experiments we also replace DTW with Euclidean distance to demonstrate that DTW is really necessary. In TABLE 10, we show a comparison of the clustering performance of TADPole to some well-known state-of-the-art clustering algorithms (which we carefully tuned) under DTW in five randomly chosen datasets from [25]. As we can see, the cluster quality returned by TADPole is usually better than the best-performing clustering algorithm. Note that we are not claiming DP is always superior, rather we chose DP because it is *at least* competitive with the state-of-the-art, and amiable to acceleration as we have demonstrated.

The greatly superior accuracy of TADPole makes the timing results somewhat irrelevant, but TADPole is at least an order of magnitude faster than the rival methods (exact numbers at [64]).

TABLE 10
CLUSTERING QUALITY (IN TERMS OF RAND INDEX) OF TADPOLE VS. SOME STATE-OF-THE-ART CLUSTERING ALGORITHMS

| Dataset | TADPole_DTW (*TADPole_ED*) | k-means[21] DTW_version | Hierarchical DTW_version | DBSCAN [12] DTW_version | Spectral [36] DTW_version |
|---|---|---|---|---|---|
| CBF | 1 (*0.66*) | 0.78 | 0.73 | 0.77 | 0.76 |
| FacesUCR | 0.92 (*0.86*) | 0.87 | 0.85 | 0.77 | 0.94 |
| MedicalImages | 0.66 (*0.67*) | 0.67 | 0.62 | 0.65 | 0.69 |
| Symbols | 0.98 (*0.81*) | 0.93 | 0.78 | 0.91 | 0.95 |
| uWaveGesture_Z | 0.86 (*0.84*) | 0.85 | 0.83 | 0.8 | 0.86 |

In addition to the comparisons mentioned above, we compare against a recent partitional clustering algorithm called k-Shape [37]. The k-Shape algorithm considers the shapes of time series by using a normalized version of the cross-correlation measure. Depending on the properties of the distance measure, this algorithm computes the cluster centroids, which are used to update the assignment of objects to cluster centers in an iterative fashion. In order to demonstrate the robustness of k-Shape, the authors compare the clustering time and quality against a large number of state-of-the-art clustering algorithms. According to these experiments, k-Shape outperforms its rivals in terms of both clustering quality and time.

We compare the cluster quality and time of TADPole against k-Shape. We exhaustively run experiments on all the datasets chosen by the k-Shape authors. To be fair, we use the same data folds for both these algorithms. For TADPole, we use a fixed 5% warping window for consistency. We show the cluster quality result in Fig. 11.

As we can see from Fig. 11, for the majority of the datasets, TADPole does much better than k-Shape. It is only in the ECGFiveDays and SonyAIBORobotSurface datasets, that k-Shape performs significantly better than TADPole.

Since these two datasets are outliers, it is worth considering why they seem to favor k-Shape.

In the case of the SonyAIBORobotSurface dataset, the start and end time of the gaits are not aligned well in terms of time, which means that the first and last points of the two time series being compared may have very different values. In Fig. 12, we show two example time series from the two classes of this dataset.

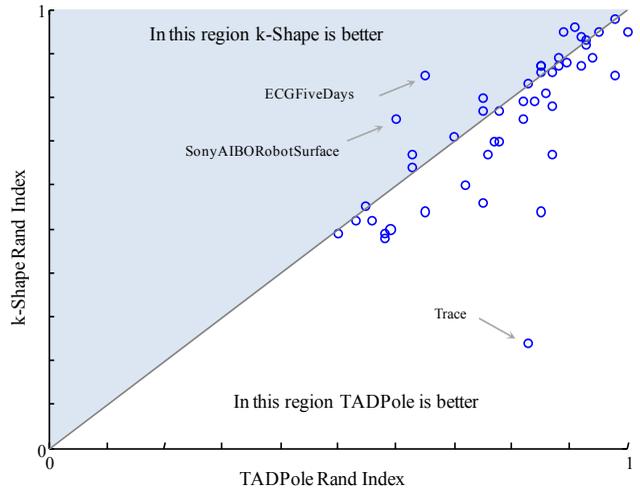

Fig. 11. Cluster quality comparison experiment of TADPole against k-Shape in terms of Rand Index. For most of the datasets, TADPole gives significantly better clusters than k-Shape.

As we can see in Fig. 12, the start points of these two time series are not aligned at all, even though the end points almost agree. Because of DTW's endpoint constraint [24][44], these are forced to match, resulting in large, near random contribution to the overall DTW distance calculation. This issue (and several solutions) has been known for decades (see Section 3 of "Endpoint Variants of the DTW Algorithm" of [39]). Rather than using one of the fixes in [39], we simply smoothed the data, which resulted in a Rand Index of 0.91 by TADPole, moving it from the "*in this region k-Shape is better*" firmly into the "*in this region TADPole is better*" camp.

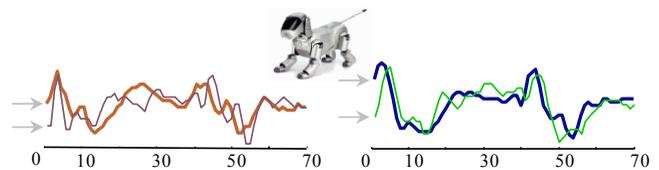

Fig. 12. *left* and *right*) Two sample individual gaits of the two classes in the SonyAIBORobotSurface dataset. The intra-class variability is poorly captured by DTW because of the great difference of the first and last endpoints of two time series under question (marked with arrows).

A similar observation explains results with the ECGFiveDays dataset, because the objects were extracted by an imperfect beat extractor, and therefore the data endpoints are not perfectly aligned (cf. Fig. 13.*left*). In addition to this, the disagreement of the start and last endpoints of the data is present (cf. Fig. 13.*right*) in this dataset. Once again, off-the-shelf smoothing (Matlab's smooth function with the default parameter) is enough to mitigate most of the problem. We ran TADPole on the smoothed data and the improved Rand Index for this dataset was 0.84, which is very close the k-Shape clustering

(0.85).

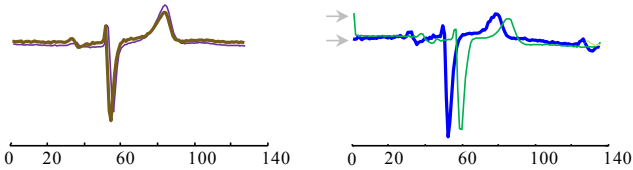

Fig. 13. *left* and *right*) The individual heartbeats of the two classes in the ECGFiveDays dataset. *left*)The heartbeats are not perfectly aligned because of an imperfect beat extractor. *right* )Like the SonyAIBORobotSurface dataset, this data also has the disagreement of the first and last endpoints (shown with arrows).

Finally, another recently published time series clustering technique called YADING is shown to "*provide theoretical proof which... ...guarantees YADING's high performance*" [10]. However, these guarantees are *only* with respect to Euclidean distance. The only publicly available real dataset the authors of [10] test on is StarLightCurves, where they obtain a Normalized Mutual Information (NMI) score of 0.60. However, TADPole can achieve an NMI of 0.61, which is very slightly better. Likewise, in an expanded tech report that augments the paper [11], the method achieves an NMI of 0.61 on the CinC_ECG_torso dataset and 0.74 on MALLAT dataset, whereas TADPOLE achieves NMIs of 0.76 and 0.84, which are significantly better.

## 5.2 Parameter Sensitivity Experiments
To demonstrate that TADPole is not sensitive to parameter choices, we took the *Symbols* and *Insect* dataset [25] and performed TADPole on it with $k = 6$ and $k = 11$ respectively. We then varied the cutoff distance ($d_c$) parameter and measured the Rand Index obtained with the alternative settings. As we can see from Fig. 14. (a) and (b), there is a very wide range of choices for the values of $d_c$, which leads to high-quality clustering. We did the same experiment for several other datasets, (details available in [64]) and the results confirm our claim that TADPole is not sensitive to this cutoff distance parameter.

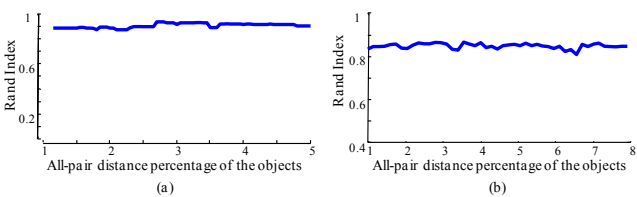

Fig. 14. (a) and (b): A parameter sensitivity test of TADPole shows stability of clustering over a very wide range of parameters.

In spite of this finding, it is clearly desirable to have some guidance in parameter setting. In the next section we introduce such a heuristic.

## 5.3 Heuristic for Setting Parameters
Up to this point we have glossed over the issue of setting the values of the threshold and the warping window width, we will now repair that omission.

First, we note that this problem is not unique to our setting, most unsupervised algorithms have this challenge. Our first proposed solution is therefore the default idea in the community. For any given problem, we can simply find the most similar dataset for which we do have labels, and hope that the best settings there (discovered by cross validation) will generalize to the current setting.

However, beyond this we do have an idea which allows us to find good (not necessarily optimal) parameter settings in most cases. The idea is very simple, we use our unlabeled dataset at hand to build a new dataset for which we do have some labels, and use this labeled dataset to do cross validation, to set the parameters.

This idea leaves open the question of where we can find some class labeled data, our solution is to *make it*. Our basic idea is simple. Before we perform any clustering, we randomly sample objects from the dataset, which we call set R. For each object O in R, we create a copy of it that we denote as Ō. We add some warping to Ō, and place it into the dataset with the same pseudo-class label as O. The intuition is that because we know that object Ō is just a minor variant of O, we can safely assume that had Ō occurred naturally, it would have been in the same cluster as O, and our must-be-in-same cluster constraint was warranted. We denote the set containing all such warped version of objects in R, R̄.

At this point, for all object Ō in R̄, we know a pseudo label. Given this, in addition to the must-be-in-same-cluster constraint, we can say that if an object in R has a different label than a pseudo label of a new object Ō, then for this pair, the must-not-be-in-same-cluster constraint can be warranted.

This idea seems to have a tautological paradox to it. It seems that if we add $w$ amount of warping to the dataset, we will discover $w$ warping in that dataset. However, this is not the case. A good setting for $w$ depends not only on the intrinsic variability of the time axis and on the size of the dataset, but on the time series shapes themselves.

Having these two constraints in hand, we design a scoring function that measures the number of object pairs in <R, R̄ > satisfying these constraints for different warping windows over all possible pairs. Therefore, our score function is:

$$score = \frac{number\ of\ pairs\ in\ <R,\bar{R}>satisfying\ link\ constraints\ for\ different\ warping\ window}{all\ possible\ pairs\ of\ <R,\bar{R}>} \quad (1)$$

We note the following:
• This idea, of generating new data, by randomly perturbing real data, is not novel [8][53] in general, but appears to be novel in the domain of time series.
• We are not claiming that this idea is the final or best solution to this issue. We merely introduce it as an existence proof that it is possible to set the parameters to reasonable settings, even in the absence of labeled data.

For some randomly chosen datasets, we compare the Rand Index obtained by TADPOLE for different warping window sizes against the score from equation 1. Our intuition is, if the shape of the score curve agrees closely with the Rand Index curve, then this score function has good correlation with the Rand Index. Having set a reasonable value of $d_c$, for warping window sizes where this score is relatively 'higher', we assume those warping window sizes as 'reasonable' parameter values. We illustrate our results in Fig. 15.

From Fig. 15 we can see that the score obtained corresponding to different warping window widths matches closely with the associated Rand Index for a fixed $d_c$. As long as the warping window width stays in relatively high scoring region, we can consider the parameter choice good.

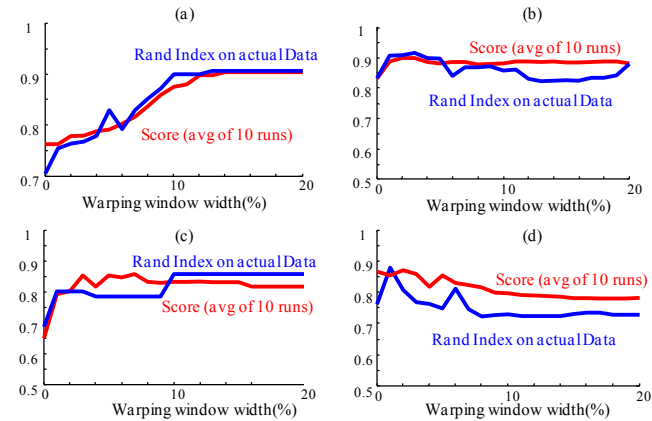

Fig. 15. Warping window width vs. Rand Index (blue) and vs. score (red) plots for Trace (a), FaceAll (b), FaceFour (c) and SwedishLeaf (d) datasets [25]. The shapes of the red and blue curves agree closely, indicating good correlation. For a fixed $d_c$, the regions of the red curve with relatively high values are indicative of good parameter (here warping window width) values.

At this point we are now able to address the following question: *given a reasonably well set warping window width, how to estimate $d_c$?* We take the same approach to set $d_c$ as we did for setting the warping window size. First we set a reasonably well value of the warping window size for a fixed $d_c$ according to the scoring function described above. Now we fix the warping window width to any value in the 'good parameter zone' and change the value of $d_c$. To quantify how well our selection of $d_c$ is, we consider the same scoring function we used for the warping window. Just like what we did for setting the warping window, we vary $d_c$ and record the score associated. The range of values resulting relatively high scores is the good parameter zone for $d_c$. For the four datasets shown in Fig. 15, by keeping the warping window size fixed, we find the best $d_c$ values listed as in TABLE 11.

TABLE 11
BEST VALUE OF DC FOR REASONABLY GOOD WARPING WINDOW WIDTH

| Dataset | Warping Window Width | $d_c$ |
|---|---|---|
| Trace | 14% | 1.7 |
| FaceAll | 2% | 3.25 |
| FaceFour | 6% | 4.8 |
| SwedishLeaf | 2% | 0.65 |

## 6  CASE STUDIES

### 6.1  Electromagnetic Articulograph Dataset

The Electromagnetic Articulograph (EMA) is a device that is increasingly used for mouth movement studies [55]. The apparatus consists of a set of unobtrusive accelerometers that are attached to multiple positions on the tongue, lips, jaw, nose and forehead, and can record high-resolution 3D movement/position data in real-time. Recent research has suggested that articulographs may eventually allow a "silent speech" interface that translates non-audio articulatory data to speech output, an idea that has significant potential for facilitating oral communication after a laryngectomy [55]. The most common use of articulographs is in speech therapy for a plethora of speech disorders. However, there is a significant obstacle to EMA use: setting up the system can take up to 30 minutes per session (this time is spent carefully gluing the sensors to the participant's face and tongue). Given this significant setup time, practitioners are anxious to get the most from each session, yet the goals of the session are not typically fixed, but rather change in reaction to the participant's progress and areas of difficulty. Thus, there is a need to cluster the utterances of speakers in an *interactive* fashion, so that sessions can be adapted on the fly. We consider a dataset of lower-lip accelerometer time series movement data of 18 words collected from multiple speakers, for a total of 414 objects. The duration of utterance of each of the words is ~0.7 seconds. In Fig. 16.*left*), we show the data collection process for one of our subjects. In Fig. 16.*right*), we show two examples of the utterances of the word "*fate*" by two different individuals. These examples are clearly warped, suggesting this is an appropriate domain for TADPole.

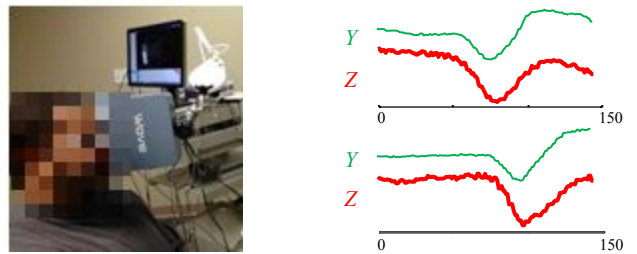

Fig. 16. *left*) One of our volunteers wearing the articulograph apparatus. *right*) Two examples of the 3D time series produced by enunciating the word "*fate*" show inter-subject warping. The X axis is omitted here, as it only has useful information for patients with facial asymmetry.

We tested the power set of combinations of axes, confirming that axis Y gives us the best clustering, with a Rand Index of 0.94 (the other results are archived at [64]). We can now ask how effective our pruning strategy is when performing this clustering. We show the result in Fig. 17.

As we can see from Fig. 17, TADPole achieves ~94% pruning, converging almost as quickly as the oracle algorithm. In wall clock terms, TADPole takes only 2.89 seconds, which is fast enough to provide interactive analysis and feedback to the patient.

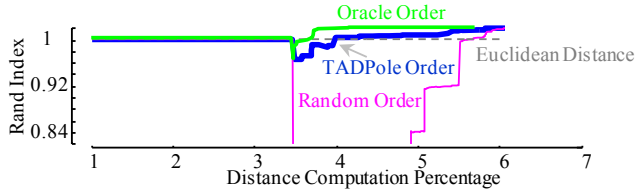

Fig. 17. The amount of distance pruning achieved by TADPole is ~94% for the Articulographs [55]. Moreover, the ordering heuristic is almost as good as the oracle ordering.

## 6.2 Pulsus Dataset

Pulsus Paradoxus is defined as a significant decline in the pulse with inspiration. It is a symptom of *Cardiac Tamponade*, a life-threatening condition where high-pressure fluid fills the sac surrounding the heart, impairing cardiac filling and causing 20,000 deaths per year in the USA alone. Of the several ways to detect Pulsus, the least invasive and simplest uses the PPG (PhotoPlethysmoGram) shown in Fig. 18.*top*).

For this case study we consider a dataset of 500 PPGs from two sources: the MIMIC II Waveform Database [16][46] and our collaborators. The latter dataset has the advantage that our collaborators followed up on the patients (in some case, *post-mortem*); thus, we have access to unusually rich annotations and external knowledge to evaluate our result. As shown in Fig. 18.*top*), the raw PPG data is very complex, and following the suggestions of Dr. John Criley (UCLA School of Medicine), we converted the PPGs to *amplitude spectrums* (Fig. 18.*middle*) and clustered in that space. Dr. Criley's intuition is that for Tamponade patients, the fluid that fills the sac surrounding the heart will cause a "shadow" signal to show up during respiration.

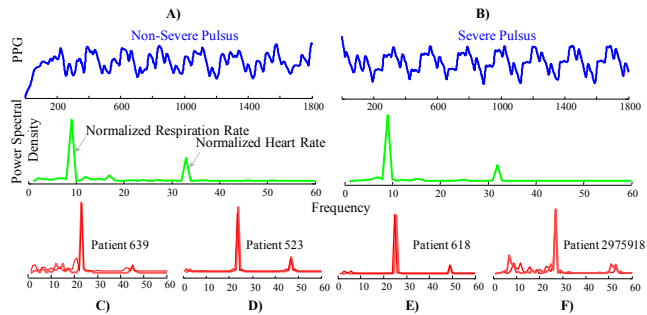

Fig. 18. *top*) PPG and Power Spectral Density (PSD) signal of a patient with non-severe Pulsus (*top.left*) and severe Pulsus (*top.right*)). *bottom*) Four PSDs of four patients forming four different clusters within the non-Pulsus objects. From these four clusters, we can see the objects are clearly warped.

For this dataset, TADPole produced a perfect clustering with a pruning rate of 88%, making it an order of magnitude faster than brute-force.  In Fig. 19.*left*), we show the PPG measurement process. From Fig. 19.*right*), we can see that the Pulsus instances are within a compact cluster and the non-Pulsus instances seem to form a number of sub-clusters. Our medical collaborator suggests this reflects the fact that there is *one way* to have Tamponade, but *multiple* ways to have a healthy heartbeat.

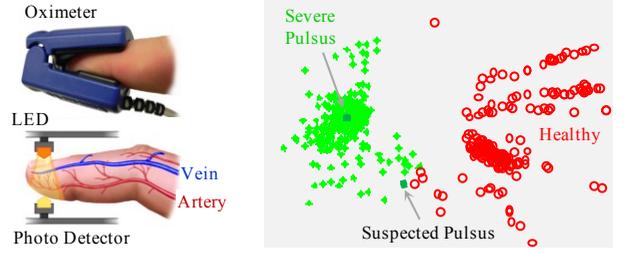

Fig. 19. *left*) A PPG apparatus. *right*) The Pulsus dataset projected into two dimensions using multidimensional scaling, and color coded by the output of TADPole.

## 6.3 Person Re-Identification Dataset

Person re-identification is the task of recognizing individuals across spatially disjointed cameras [15], and is an important problem for understanding human behavior in areas covered by surveillance cameras. As shown in Fig. 20, we can extract color histograms from the video, thus treating the problem as a multidimensional time series problem. We considered the PRID dataset [20], randomly extracting 1,000 images of 12 different individuals.  After we ran TADPole in this dataset to cluster the images of these 12 individuals, we achieved a Rand Index of 95.4% and distance pruning of 80% (*anytime* plot at [64]). In contrast, Euclidean distance only achieves a Rand Index of 89%.

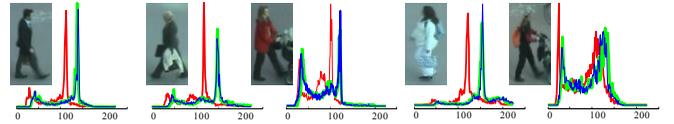

Fig. 20. Representative images from the dataset [20] with their corresponding color histograms.

## 6.4 Clustering Multidimensional Data

In order to demonstrate TADPole's suitability for multidimensional data, we clustered two real-world datasets - Cricket and uWaveGesture.

### 6.4.1   Cricket Dataset

Cricket is a very popular game around the globe. As part of the refereeing, an umpire uses a fixed set of gestures using his hands and (sometimes) legs to communicate his decisions to a distant scorekeeper. The interested reader may find a list of complete signals in [32].

In this case study, we analyze the utility of TADPole for gesture recognition applications. For this, we use the dataset in [7], where 4 different umpires perform the various signals. On average, each signal is performed 16 times. Two accelerometers were attached to the wrists of the umpires and the data was sampled at 184 Hz. Considering both the left and right hands, this data has 6 dimensions (each accelerometer has X, Y and Z components).

Fig. 21 shows what each of the seven gestures *Last Hour, Leg Bye, No Ball, One Short, Out, Penalty Runs* and *Six* look like.

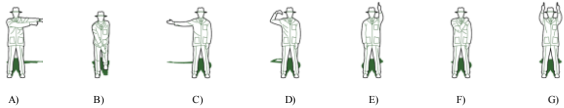

Fig. 21. The seven cricket game gestures we used for our experiment. A) Last Hour, B) Leg Bye C) No Ball, D) One Short, E) Out, F) Penalty Runs and G) Six.

In this data setting, we picked the tri-axial right hand signals as the three dimensions we wish to cluster. The number of objects in this dataset is 258 with each of the time series being of length 1,155. We used warping window = 5%.

In this dataset, TADPole gives a Rand Index of 0.92 with ~82% pruning. In wall clock terms, TADPole takes only 52 seconds to cluster this 27 minutes long data. That is to say, we can cluster the data about 30 times faster than real time.

### 6.4.2 uWaveGestureLibrary Dataset

For this experiment we consider the three dimensional uWaveGestureLibrary dataset [25]. This data was collected to support efficient personalized gesture recognition on a wide range of electronic and mobile devices. This dataset contains eight simple gestures identified by a Nokia research study. The gestures were collected from eight participants with the Wii remote-based prototype. Fig. 22 shows these gestures as the paths of hand movement [31][38].

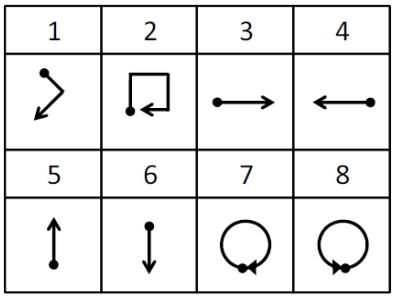

Fig. 22. The gesture vocabulary adopted from [31][38].

In this experiment, as before, we are interested in the actionability of TADPole to cluster human gestures. We use all eight gestures as input to TADPole. The number of objects in this dataset is 896 with each of the time series being of length 315 and a sampling rate of 100 Hz. After we run TADPole on this data, we achieve a Rand Index of 0.93 with ~88% pruning. In wall clock terms, TADPole needs only 56 seconds to cluster this 47 minutes long data. As before, we can cluster the data much faster than real time.

### 6.5 Generalizing TADPole to Cluster Discrete Data

Until now, we have shown that our TADPole framework can efficiently perform exact clustering of *real-valued* time series data. The interested reader might wonder whether this framework is general enough to be extended for handling other types of data. In this section we present a dis-cussion of the generalization of this framework to cluster very long *discrete* biological sequences. We note here that our experiments are not claiming a novel discovery of any biological significance. We are merely demonstrating a "proof-of-concept," that the TADPole framework can be extended to perform fast clustering of long biological strings.

More generally, we believe that this example shows that the TADPole framework may be useful in scaling up clustering for *any* distance measure for which one can compute both upper and lower bounds. Beyond DTW, this would include most string similarity measures (as we show below), the Earth Movers Distance, Graph Edit Distance etc.

### 6.5.1 Clustering of Protein Sequences with Edit Distance

The identification of biological sequences with similar functionality requires similarity search in large databases, and is a well-known problem in the Bioinformatics community [14][19]. As an example, there exist diseases such as Maple Syrup Urine, Propionic Acidemia, Hurler Syndrome [34], etc. which are caused by the lack of sections of protein sequences in an organism, and can be life-threatening. In order to design drugs for such diseases, biologists produce sequences with the same functionality, and insert these sequences into the organisms to compensate the deficit.

Typically these protein strings are very long, therefore similarity search in such a large sequence database can be computationally challenging [2]. For example, large genomes are usually expressed using Expressed Sequence Tag (EST) databases. In such databases, gene portions are represented by mature mRNA, which typically are 500-800 nucleotides long. Therefore, similarity-based algorithms in such biological databases require algorithms that can handle large strings efficiently. We see TADPole as a potential general framework for efficiently clustering such biological data.

In order to measure the similarity between two sequences of biological strings, researchers often use Edit Distance (EdD), or one of its many variants [29]. Given two strings A and B, the EdD is defined as the minimum number of deletion(s), insertion(s), and substitution(s) needed to transform A into B. We list the notation of EdD in TABLE 12:

TABLE 12

Symbol Table

| Symbol | Explanation |
|---|---|
| α | Finite alphabet |
| α* | Set of all finite sequences of symbols from α |
| *a,b* | Variables presenting individual symbols |
| A,B | Variables denoting sequences from α* |
| \|A\| | Length of sequence A |
| A(i,j) | Subsequence from the $i$th to the $j$th symbol in A |
| *a*[i] | $i$th symbol in sequence A |

Given the notations above, a substitution operation $S(a[i], b[j])$ is defined as replacing $a[i]$ in A by $b[j]$ in B. An insertion operation $I(i, b[j])$ is inserting $b[j]$ in the $i$th location of A. Deleting the symbol $a[i]$ in A is denoted by $D(a[i], -)$. Each of these edit operations is assigned a weight which represent the difference between two symbols or the penalty for insertion/deletion operation. A transformation from A to B, $T(A,B)$ is the set of edit operations applied to A iteratively which transform A to B. The weight of this transformation, $w(T(A,B))$ is defined as the sum of the weights of the operations in $T(A,B)$. Given these definitions, the edit distance between A and B, $EdD(A,B)$ is defined as, $EdD(A,B) = \min(w(t))$, $\forall t \in T(A,B)$.

As an example, for two strings A = 'INDUSTRY' and B = 'INTEREST', the optimal set of editing operations for $T(A,B)$ are:

Starting string = INDUSTRY, Goal String = INTEREST
$S(A[3],B[3]) = IN\underline{T}(D)USTRY$
$S(A[4],B[4]) = INTE(U)STRY$
$I(5,B[5]) = INTE\underline{R}STRY$
$I(6,B[6]) = INTERE\underline{S}TRY$
$D(A[9],-) = INTEREST-(R)Y$
$D(A[9],-) = INTEREST-(Y)$

Therefore $EdD(A,B) = 6$.

Using dynamic programming, $EdD(A,B)$ can be solved in $O(nm)$ time and $O(n+m)$ space, where $n$ and $m$ are the lengths of A and B respectively. For very large biological sequences, the computation of edit distance can therefore be very time consuming. We propose to exploit our TADPole framework to accelerate edit distance based clustering of long biological strings. Both edit distance and DTW are elastic distance measures [9] and for both of these measures, tight lower and upper bounds are already defined. Therefore, our TADPole framework is a suitable fit to perform fast and exact clustering of long biological strings. For clarity however, we confine our discussion only to the clustering of protein sequences.

We use Edit Distance (EdD) to perform the clustering of long protein sequences with TADPole framework. According to the framework structure, the inputs are the fully computed lower and upper bound matrices of the edit distance of all the objects of the dataset. As the lower bound of the edit distance, we use a notion of the distance that maps the strings of the database onto a multidimensional integer space using a wavelet based method [22]. We use the difference of the length of the strings in the dataset as the upper bound of the edit distance. Tighter bounds are known, but we are only interested in the general principle here.

In order to demonstrate the utility of TADPole in accelerating the clustering of large protein sequences, we use a random chunk of $^{1600}_{801}$ in [27] of the UniProt dataset [54]. It includes 178 strings defined over an alphabet of 21 letters. The protein strings have variable lengths ranging from 1052 to 1597. We cluster this dataset using TADPole and obtain distance pruning of 28%. Although the speedup we achieve does not look impressive at the first sight, we believe that with the application of tighter upper and lower bound of edit distance, TADPole would further improve.

Given the distance pruning we achieve, it is instructive to see the utility of the clusters returned by TADPole. Out of the eight clusters returned by TADPole, one represents the rpoB gene which encodes the beta subunit of bacterial RNA polymerase. According to the STRING database, the DNA-dependent RNA polymerase "...*catalyzes the transcription of DNA into RNA using the four ribonucleoside triphosphates as substrates.*" [51] Another cluster corresponded to cell division protein mukB, which "...*plays a central role in chromosome condensation, segregation and cell cycle progression .*"[51]

Considering the fact that the clusters returned by TADPole have some level of biological significance, the readers might wonder how fast the algorithm converges to the ground truth. We show the result in Fig. 23.

From Fig. 23 we can see that even if the amount of pruning achieved by TADPole is not impressive at the very first sight, the quality of the clusters in the earlier stages of the algorithm is *very* impressive, and it is almost as good as the oracle ordering. Therefore, with very small amount of actual edit distance computation, TADPole can produce very high quality cluster results.

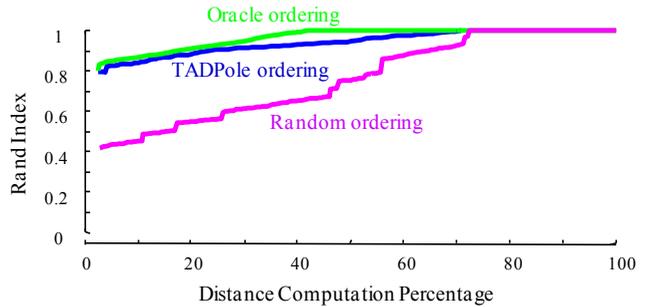

Fig. 23. The fast convergence of TADPole ordering (blue) to the oracle ordering (green) shows that the quality of the clusters in the earlier iterations is quite good. The cluster quality given by the random ordering (pink), however, is not as good as TADPole.

## 7   CONCLUSIONS

By introducing novel pruning strategies that exploit *both* upper and lower bounds, we have produced a robust DTW clustering algorithm that is both *absolutely* much faster than the state-of-the-art algorithms, and able to compute the clustering in an *anytime* fashion. We have demonstrated the utility of our algorithms on diverse domains, including two in which our algorithm is currently actively deployed (EMA and Pulsus). We have made all our code and data publicly available to allow the community to exploit and extend our ideas.

## 8   ACKNOWLEDGEMENTS

We gratefully acknowledge the financial support for our research provided by NSF IIS-1161997 II, NIH R03 DC013990, NIH R01 DC013547 and all the donors of the datasets. We further acknowledge Dr. John Criley (UCLA School of Medicine) for donating the PPG data and providing detailed explanations of the intricacies of Pulsus Paradoxus.

## 9 REFERERENCES

## 10  APPENDIX

Proof of Exactness of the TADPole Algorithm

**Theorem 1.** *The cluster result obtained by TADPole is exactly the same as the one obtained by DPDTW.*

**Proof.** In order to prove this theorem, we must first consider the following two lemmas:

**Lemma 1.** *The pruned distance computations by TADPole during the density calculation have no effect on the local densities of the objects in the dataset.*

**Proof (by contradiction).** We prove this lemma in the context of the three cases of distance computation pruning during the local density calculation.

**Case A:** *Objects i and j are identical*
When $i$ and $j$ are identical, then their actual distance $D_{ij}$ is zero, which means they are definitely within $d_c$. Therefore, we can safely prune their distance computation.

**Case B:** $UB_{Matrix}(i,j) < d_c$
In this case our claim is, $D_{ij} <= d_c$ and we can safely prune the distance computation. For the sake of the proof, let us assume that $D_{ij} > d_c$.
By definition, $LB_{Matrix}(i,j) <= D_{ij} <= UB_{Matrix}(i,j)$
Therefore, because $UB_{Matrix}(i,j) < d_c$, and $D_{ij} <= UB_{Matrix}(i,j)$, $D_{ij} \not> d_c$. Therefore, our claim holds.

**Case C:** $LB_{Matrix}(i,j) > d_c$
In this case our claim is, $D_{ij} > d_c$ and we can safely prune the distance computation. For the sake of the proof, let us assume that $D_{ij} <= d_c$.
By definition, $LB_{Matrix}(i,j) <= D_{ij} <= UB_{Matrix}(i,j)$
Therefore, because $LB_{Matrix}(i,j) > d_c$, and $D_{ij} >= LB_{Matrix}(i,j)$, $D_{ij} \not<= d_c$. Therefore, our claim holds.
TADPole only calculate the distances for the case when $LB_{Matrix}(i,j) < d_c$ and $UB_{Matrix}(i,j) > d_c$. Therefore, lemma 1 holds.□

**Lemma 2.** *The pruned distance computations by TADPole do not affect the NN distance of any object calculation from its higher density list.*

**Proof (by contradiction).** We prove this lemma in the context of the two phase pruning during NN distance calculation of an object from its higher density list.

**Phase 1: Upper bound calculation**

For this phase we need to prove that the NN distance from its higher density list of an object $i$, $\delta_i <= ub_i$, where $ub_i$ is the upper bound of this NN distance we intend to find.

For the sake of the proof, let us assume $\delta_i > ub_i$.

Assume that the higher density list of $i$ consists of $j_{1...}j_k$.

Assume that the NN of $i$ from its higher density list is $j_n$, $1<=n<=k$. Therefore, $D_{ij_n} = \delta_i$

By definition, $ub_i = D_{ij_p}$ or , $ub_i = UB_{Matrix}(i,j_p)$ ,where $1 <= p <= k$

Therefore, $\delta_i = D_{ij_{p'}}$ when $n = p$ or $\delta_i <= UB_{Matrix}(i,j_p)$, when $n \neq k$.

From this, we can say that $\delta_i \not> ub_i$. Therefore, our claim holds.

**Phase 2: Pruning**

For this phase we need to prove that $\forall\ j \in \delta\_list_i(j))$, if $LB_{Matrix}(i, j) > ub_i$, then $j$ is not the NN of $i$ from its higher density list.

For the sake of the proof let us assume $j$ is the NN of $i$ from its higher density list.

From the proof in phase 1, $D_{ij} <= ub_i$. By definition, $LB_{Matrix}(i,j) <= D_{ij} <= UB_{Matrix}(i,j)$

But if $LB_{Matrix}(i, j) > ub_i$, then $D_{ij} > ub_i$. Therefore, $D_{ij} > ub_i$. It means that $j$ cannot be the NN of $i$. (Proved)

TADPole only prunes the computation of all $D_{ij}$ where $j$ cannot be the NN of $i$ from its higher density list. Therefore, lemma 2 holds. $\square$

The proof of lemma 1 and 2 states the exactness of our pruning strategy. Therefore, theorem 1 holds. $\square$

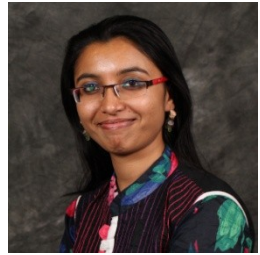

**Nurjahan Begum** obtained her PhD in Computer Science from University of California, Riverside in 2016 with focus on Data Mining, Time Series Analysis and Machine Learning. Nurjahan has over 10 publications in top data mining and machine learning venues including SIGKDD, VLDB, ICDM, SDM, CIKM, etc. Currently Nurjahan is a Software Engineer at Teradata Labs, where she is actively building new development enhancement features for the Teradata query optimization engine. Prior to joining Teradata, Nurjahan worked as a Research Intern in Bell Labs (2014) and Yahoo Research (2015).

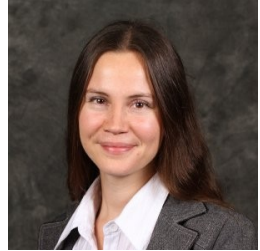

**Liudmila Ulanova** graduated with PhD in Computer Science from University of California, Riverside in Spring 2016. She was focusing on Time Series Data Mining. Liudmila coauthored over a dozen of publications including those in the top-tier data mining venues such as SIGKDD, ICDM, and SDM. After graduation Liudmila joined Analysis and Experimentation team at Microsoft Corporation as a Software Developer. She works on the Experimentation Platform which aims to accelerate software innovation through trustworthy experimentation. Before starting at Microsoft, Liudmila was a Summer Research Intern at NEC Laboratories America (Summer 2014) and Facebook Inc. (Summer 2015).

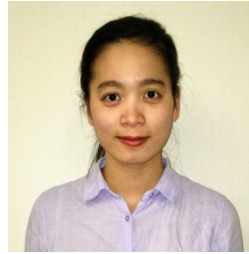

**Hoang Anh Dau** is a second-year PhD student in Computer Science at University of California, Riverside. Her research interests lie in data mining of time series. She is particularly interested in the practical applications of data mining on real world problems. She holds a Bachelor of Foreign Languages from Hanoi University, a Master of Arts in Communication and Media Studies from Monash University, and a Master of Science in Computing from RMIT University.

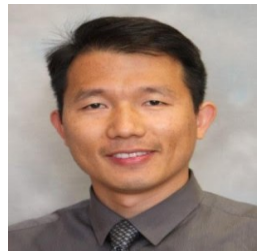

**Jun Wang** is an Assistant Professor of Biomedical Engineering and Communication Disorders at the University of Texas at Dallas. His research areas include silent speech recognition from articulatory motion, motor speech disorders due to neurological diseases, and computational neuroscience for speech. Wang earned his PhD in Computer Science from the University of Nebraska-Lincoln in 2011. He received the American Speech-Language-Hearing Foundation New Century Scholar Award in 2015. He has also authored in more than 25 papers in premier speech recognition, communication disorders, and data mining conferences and journals

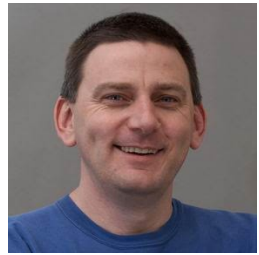

**Eamonn Keogh** is a full professor of computer science at the University of California Riverside. His research areas include data mining, machine learning and information retrieval, specializing in techniques for solving similarity and indexing problems in time-series datasets. He has authored more than 200 papers. He received the ACM SIGKDD 2012 best paper award, the IEEE ICDM 2007 best paper award, and the SIGMOD 2001 best paper award. He has given over two-dozen well-received tutorials in the premier conferences in data mining and databases.